\documentclass[letterpaper]{article} 
\usepackage{aaai2026}  
\nocopyright
\usepackage{aaai2026}
\usepackage{times}  
\usepackage{helvet}  
\usepackage{courier}  
\usepackage[hyphens]{url}  
\usepackage{graphicx} 
\usepackage{natbib}  
\usepackage{caption} 
\usepackage{algorithm}
\usepackage{multirow}
\usepackage{adjustbox}
\usepackage{enumitem}
\usepackage{amssymb}
\usepackage{algpseudocode}
\usepackage{caption}
\usepackage{amsmath}
\usepackage{diagbox}
\usepackage{booktabs}
\usepackage[table,xcdraw]{xcolor}

\usepackage{newfloat}
\usepackage{listings}
\DeclareCaptionStyle{ruled}{labelfont=normalfont,labelsep=colon,strut=off} 
\floatstyle{ruled}
\newfloat{listing}{tb}{lst}{}
\floatname{listing}{Listing}
\title{RSOD: Reliability-Guided Sonar Image Object Detection 

with Extremely Limited Labels}

\author{
    Chengzhou Li\textsuperscript{\rm 1},
    Ping Guo\textsuperscript{\rm 1},
    Guanchen Meng\textsuperscript{\rm 1},
    Qi Jia\textsuperscript{\rm 1}$^*$,
    Jinyuan Liu\textsuperscript{\rm 1},
    Zhu Liu\textsuperscript{\rm 1},\\
    Xiaokang Liu\textsuperscript{\rm 1},
    Yu Liu\textsuperscript{\rm 1},
    Zhongxuan Luo\textsuperscript{\rm 1},
    Xin Fan\textsuperscript{\rm 1}\thanks{Corresponding author.}
}
\affiliations{
    \textsuperscript{\rm 1}School of Software Technology, Dalian University of
Technology\\

    \{lichengzhou, guoping, guanchenmeng, liuzhu, xkliu\}@mail.dlut.edu.cn,
    
    \{jiaqi, liuyu882, zxluo, xin.fan\}@dlut.edu.cn, atlantis918@hotmail.com
}

\usepackage{bibentry}

\begin{document}

\maketitle

\begin{abstract}
Object detection in sonar images is a key technology in underwater detection systems. Compared to natural images, sonar images contain fewer texture details and are more susceptible to noise, making it difficult for non-experts to distinguish subtle differences between classes. This leads to their inability to provide precise annotation data for sonar images. Therefore, designing effective object detection methods for sonar images with extremely limited labels is particularly important. To address this, we propose a teacher-student framework called RSOD, which aims to fully learn the characteristics of sonar images and develop a pseudo-label strategy suitable for these images to mitigate the impact of limited labels. First, RSOD calculates a reliability score by assessing the consistency of the teacher's predictions across different views. To leverage this score, we introduce an object mixed pseudo-label method to tackle the shortage of labeled data in sonar images. Finally, we optimize the performance of the student by implementing a reliability-guided adaptive constraint. By taking full advantage of unlabeled data, the student can perform well even in situations with extremely limited labels. Notably, on the UATD dataset, our method, using only 5\% of labeled data, achieves results that can compete against those of our baseline algorithm trained on 100\% labeled data. We also collected a new dataset to provide more valuable data for research in the field of sonar.
\end{abstract}
\noindent\textbf{Code} --- https://github.com/LICZ9/RSOD

\section{Introduction}

Seawater absorbs visible light very strongly, with effective attenuation depths for different wavelengths of only a few meters to a few tens of meters~\cite{xie2022dataset}. In turbid waters, the deep sea, or night-time environments, visible-light imaging is almost entirely ineffective~\cite{liu2025dcevo,li2025difiisr}. In contrast, sonar is inherently robust to changes in illumination, suffers a much lower propagation loss, and can detect objects at ranges of tens to hundreds of meters~\cite{cao2022dynamic}. As a result, it has been widely adopted for a variety of underwater tasks, including but not limited to accident rescue, equipment maintenance, and biological surveys~\cite{cao2022dynamic,mccann2018underwater}. However, despite these advantages, sonar images remain prone to interference from electronic noise, suspended particles, and sediment, resulting in sparse and difficult to interpret object textures~\cite{singh2021marine,li2012digital}. At the same time, the high confidentiality of the sonar data and the steep cost of annotation make it difficult for supervised learning to obtain sufficient labeled data~\cite{li2025physics}. Therefore, exploring the intrinsic features of sonar image and developing efficient strategies to mitigate label scarcity are crucial to improving underwater robot object detection performance.

\begin{figure}[t]
    \centering
    \includegraphics[width=\linewidth]{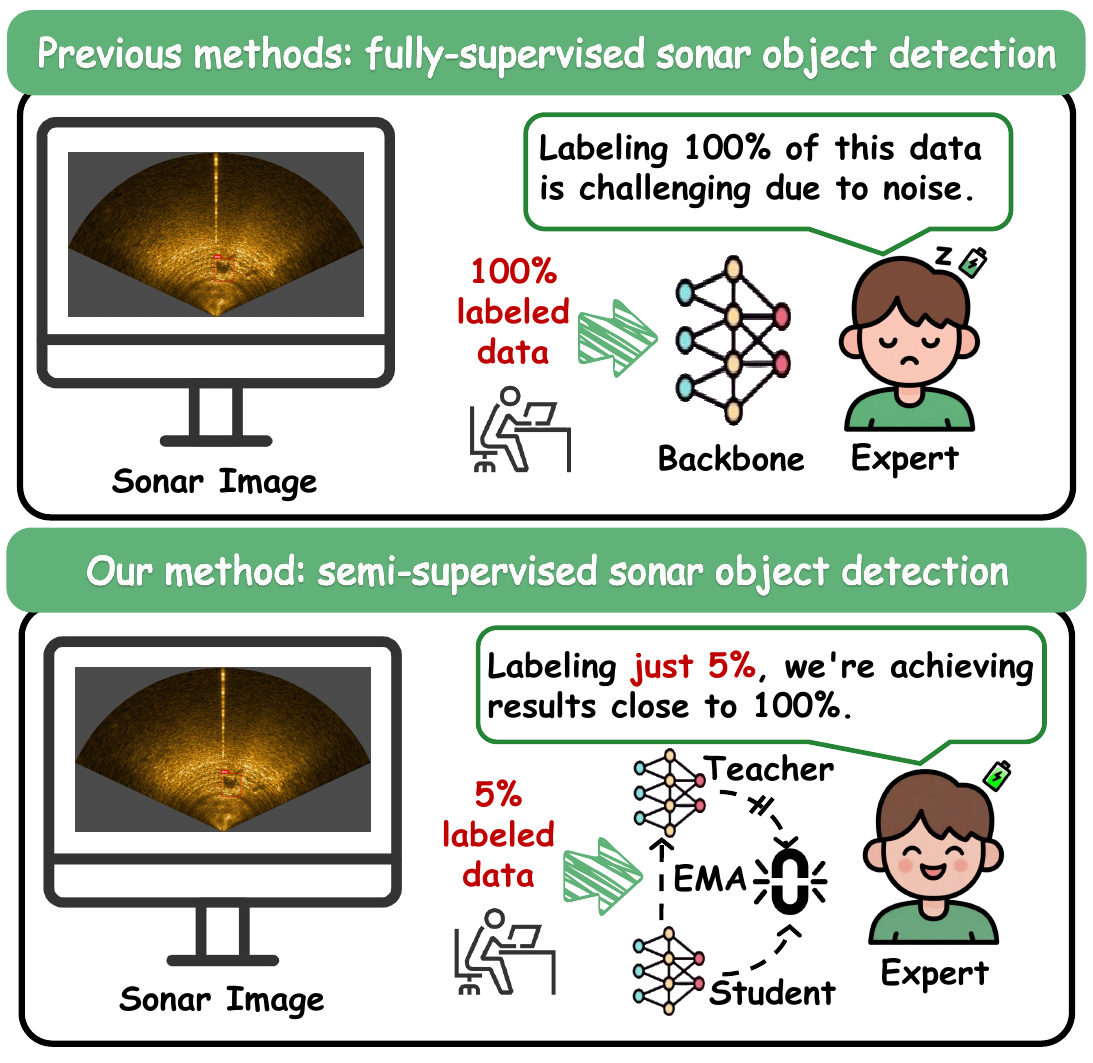}  

    \caption{Compared to previous approaches for training sonar image object detection detectors, our method reduces the requirement for labeled data.}
    \label{fig:fig1}
\end{figure}
As shown in Fig.~\ref{fig:fig1},  to address the issues above, we introduce semi-supervised learning into sonar object detection to leverage unlabeled data and reduce our dependence on manual annotations. Typical semi-supervised methods~\cite{xu2021end,liu2024two} generate pseudo-labels for unlabeled samples through model predictions, which are then used to train a student~\cite{hua2023sood}. Specifically, the teacher first applies weak augmentations (e.g. mild noise, translation, or rotation) to unlabeled sonar images, thereby generating a set of pseudo-labels~\cite{zhang2022semi}. The student then enforces prediction consistency by using these pseudo-labels to supervise strongly augmented versions of the same images (e.g. random cropping, heavy noise injection).

However, due to the typically limited texture detail information of objects in sonar images, the noise in pseudo-labels increases significantly in scenarios with extremely limited labels, severely restricting the model's stability and robustness in sonar object detection tasks. Furthermore, while pseudo-labels can significantly enhance the recognition of large objects and dominant classes, they provide very limited assistance for small objects and multi-object scenarios. Therefore, seeking effective methods to reduce pseudo-label noise and designing pseudo-label strategies that balance the detection capabilities of different scales of targets in sonar images are crucial to further improving detection performance in situations with extremely limited labels.

In this paper, we propose RSOD, a semi-supervised learning framework for sonar image object detection aimed at addressing the issue of limited labels. First, we evaluate the predictions of the teacher in multiple augmented views to calculate the reliability score for each pseudo-label bounding box. Next, we select pseudo-labels with high reliability scores and apply our object mixed pseudo-label strategy to effectively reduce the negative impact of pseudo-label noise and balance the model’s learning capability for objects of different scales. Finally, we optimize the student through reliability-guided adaptive constraint, maximizing the utilization of unlabeled data while improving the influence of high quality pseudo-labels on the model.

Our main contributions can be summarized as follows:
\begin{itemize}
\item
To the best of our knowledge, we are the first to introduce a semi-supervised object detection algorithm specifically designed for sonar images. Our proposed RSOD framework efficiently utilizes unlabeled samples to lower annotation costs while enhancing detector performance.

\item
We propose a novel method for calculating pseudo-label reliability scores along with an object mixed pseudo-label strategy to maximize the utilization of high quality pseudo-labels and minimize the impact of noisy pseudo-labels on the overall loss.

\item
We created a new dataset named the Forward-Looking Sonar Image Object Detection (FSOD) dataset, which comprises 3,929 sonar images captured with an Oculus M750d sonar, annotated in 10 distinct object categories.

\item
Extensive experiments conducted under different settings of labeled data ratios in the UATD and FSOD datasets demonstrate that RSOD outperforms existing state-of-the-art methods, particularly showing significant improvements when only 5\% of the data is annotated.
\end{itemize}

\section{Related Work}

\subsection{Sonar Image Object Detection}

In recent years, the rapid advancement of deep learning~\cite{yu2021real,redmon2018yolov3,liu2016ssd} has dramatically improved the performance of CNN-based sonar image object detection~\cite{tarvainen2017mean,valdenegro2016objectness}, especially Fast R-CNN~\cite{girshick2015fast} and Faster R-CNN~\cite{ren2015faster}. Faster R-CNN introduces a Region Proposal Network (RPN) that directly generates candidate boxes and scores their object in an end-to-end fashion from the raw input, sharing layers with both the classifier and the regressor, and, in doing so, substantially improves detection speed and accuracy. ~\cite{wang2022detection} have designed multi-scale convolution structures to enhance the network's ability to perceive small target features, thereby improving the detection performance of small targets in sonar images. ~\cite{shen2024epl} incorporate the characteristics of underwater sonar images, such as grayscale, low contrast, and noise interference, to optimize the performance of sonar object detectors.

However, these methods do not take into account the label scarcity problem that sonar images typically face in real-world scenarios. We introduce a teacher-student framework to fully leverage unlabeled data, thus mitigating the impact of insufficient labeled data on the model.

\subsection{Semi-Supervised Object Detection}
Sonar images, because of their high confidentiality and the inherent difficulty of interpreting overlapping echoes, suffer from an extreme scarcity of labeled data. Semi-supervised object detection (SSOD) seeks to address this by cleverly integrating potential object information from unlabeled images to boost detection performance~\cite{wang2025multi,liu2023ambiguity}, typically through a teacher-student paradigm: a teacher is first trained on the limited labeled set~\cite{DBLP:conf/iclr/LiuMHKCZWKV21,wang2023consistent}; It then generates pseudo-labels for the unlabeled images, which are used to jointly train the student model. To improve pseudo-label quality, researchers have incorporated consistency regularization~\cite{liu2022unbiased}, confidence threshold filtering~\cite{xu2021end}, and random data increase perturbations~\cite{chen2023mixed}. 

Existing SSOD methods cannot effectively address the issue of noisy pseudo-labels generated by the teacher due to the limited texture detail in sonar images. Therefore, we propose a new reliability-guided semi-supervised object detection method for sonar images that reevaluates all pseudo-labels using reliability scores, thereby effectively mitigating the impact of noisy pseudo-labels on the detector.
\section{Data Construction}
\subsection{Underwater Sonar Data Collection}

The scarcity of publicly available sonar image datasets significantly hampers progress in sonar object detection. This lack of resources limits the development of advanced algorithms and restricts research on the features of sonar images and the generalization of algorithms.

\begin{figure}[t]
    \centering
    \includegraphics[width=\linewidth]{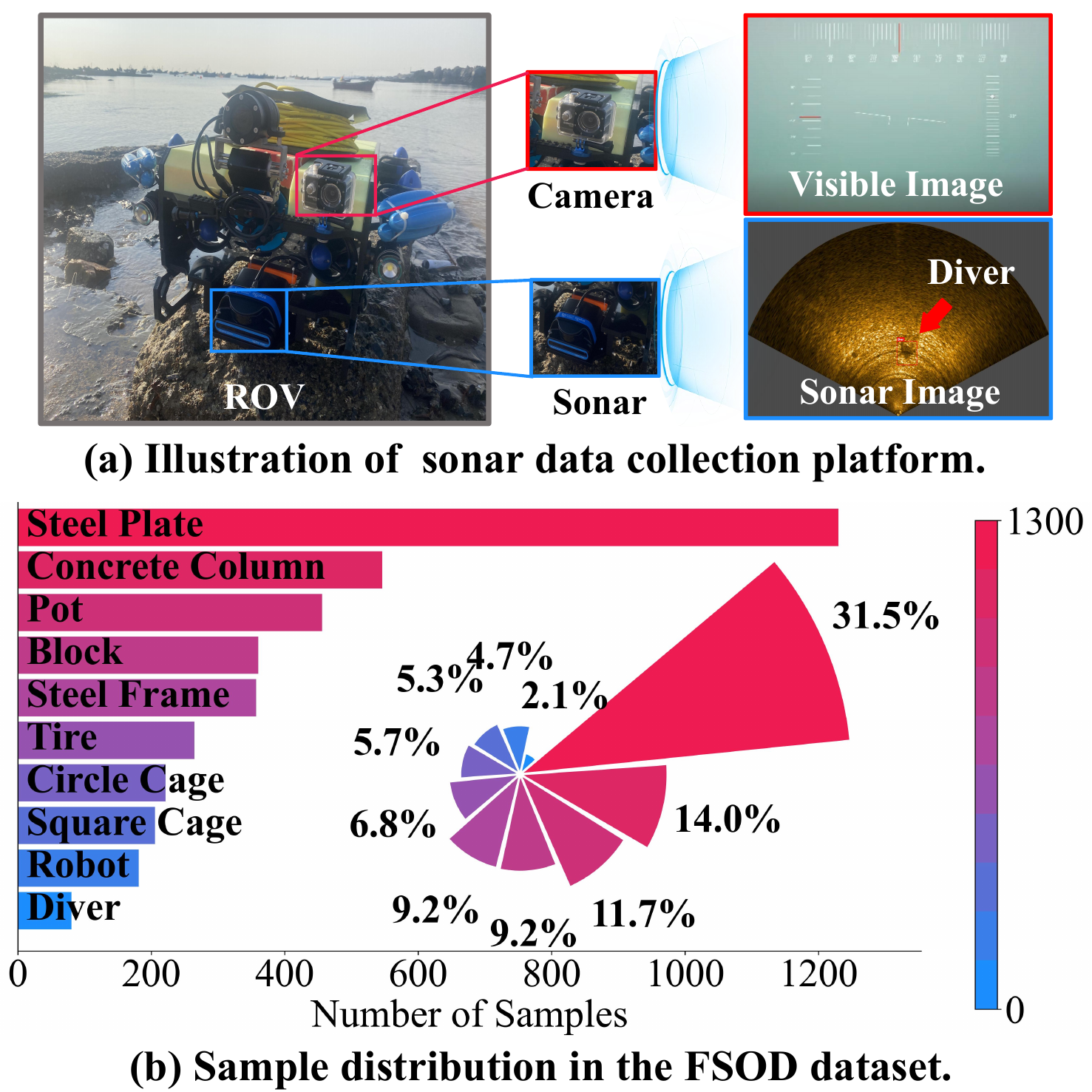}  
    \caption{(a) The collection process of the FSOD dataset. (b) The number in each category of the FSOD dataset.}
    
\label{fig:dataset_distribution}
\end{figure}

To address the aforementioned challenges, we present a new dataset, the Forward-Looking
Sonar Image Object Detection (FSOD) dataset. Data were collected in Bohai Bay, as shown in Fig.~\ref{fig:dataset_distribution}(a). Specifically, we used a remotely operated vehicle (ROV) equipped with a multibeam forward-looking sonar (Oculus M750d) to acquire underwater data. To reduce mutual interference between objects and facilitate location, the objects were connected and suspended by ropes at depths of approximately 3 to 20 meters below the water surface and placed at fixed intervals. During the data collection process, each object was imaged from different angles and distances (2–15 meters), thus enriching the diversity of the dataset. The collected objects were then filtered, pre-processed, and annotated, resulting in a dataset comprising 3,929 images across ten categories. Fig.~\ref{fig:dataset_distribution}(b) shows the number of each category in the dataset. In particular, the steel frame category contains the most samples with 1,231 images, accounting for 31.5\% of the dataset, while the diver category has the fewest with only 81 samples, accounting for 2.1\%, demonstrating the inherent long-tailed distribution of the FSOD dataset. It should be mentioned that, compared to the mannequin samples provided in the UATD dataset~\cite{xie2022dataset}, the FSOD dataset also includes a diver category, offering important data resources for the development of underwater emergency rescue, accident prevention, and diving safety assurance systems.

\begin{figure*}[t]
    \centering
    \includegraphics[width=\linewidth]{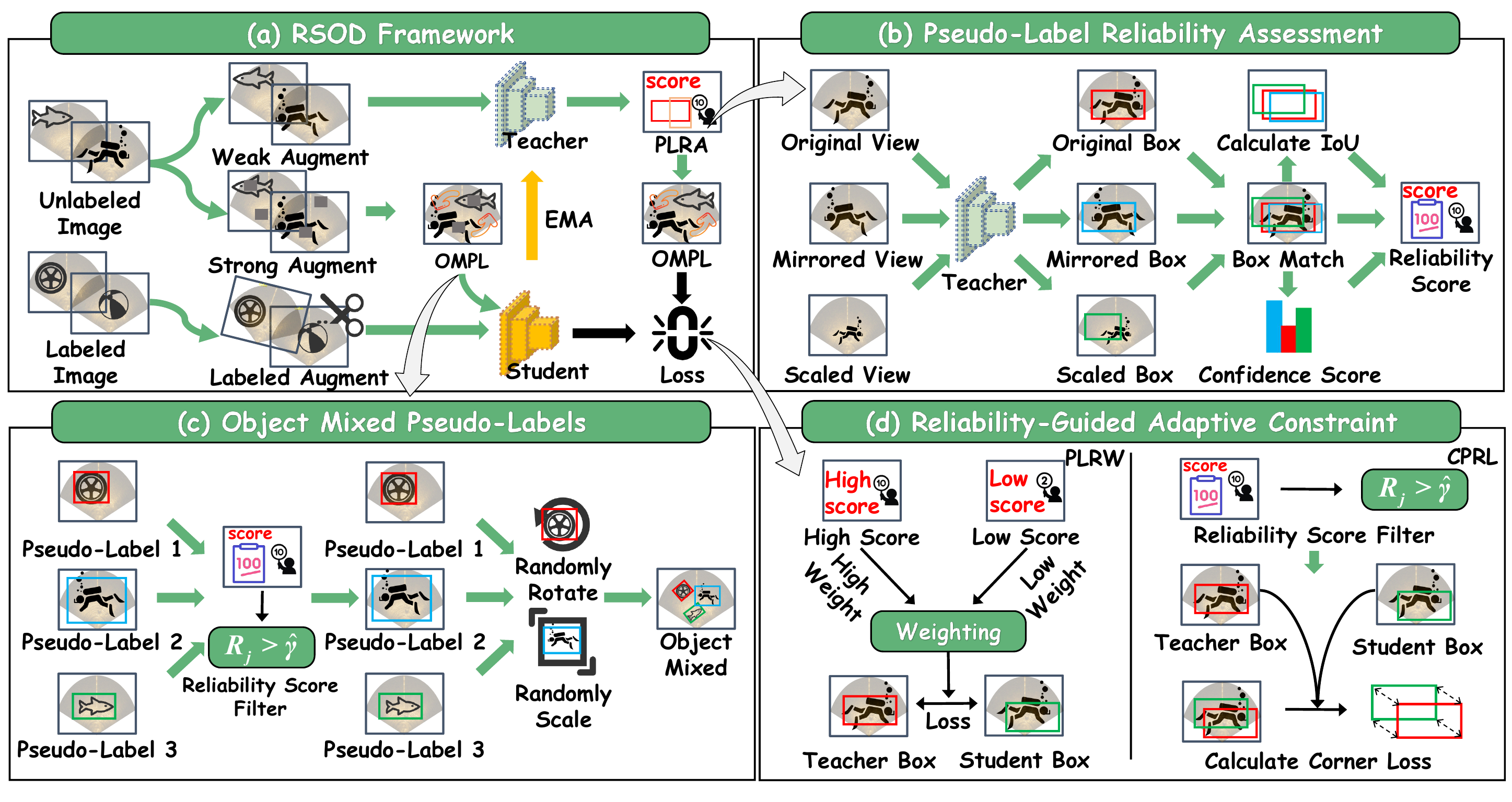}  
    \caption{The overall architecture of RSOD. (a) Illustrating the basic structure of the RSOD, (b) the process of assessing the reliability of pseudo-labels, (c) the generation process of object mixed pseudo-labels, and (d) the loss optimization process by reliability-guided adaptive constraint.}
    
\label{fig:Sonar}
\end{figure*}

\section{Methodology}
\subsection{Problem Definition}
The goal of RSOD algorithm is to learn a detector to locate and identify objects. The training dataset consists of two parts: a set of labeled sonar images $D_{l} = \{S_i^l, Y_i^l\}_{i=1}^{N_l}$ and a set of unlabeled sonar images $D_{u} = \{S_i^u\}_{i=1}^{N_u}$. Here, $N_l$ and $N_u$ denote the numbers of labeled and unlabeled sonar images, respectively. For each labeled sonar image $S^l$, the corresponding annotation $Y^l$ provides the locations, sizes, and categories of all the boxes.
\subsection{Overview}
As shown in Fig.~\ref{fig:Sonar}, the overall framework of RSOD. (a) Illustrating the basic structure of the method, (b) the process of assessing the reliability of pseudo-labels, (c) the generation process of object mixed pseudo-labels, and (d) the loss optimization process by reliability-guided adaptive constraint. In Fig.~\ref{fig:Sonar}(a), the proposed RSOD framework consists of two detection networks with identical architectures, namely the teacher network and the student network. The teacher network is responsible for generating pseudo-labels, while the student network is jointly trained using both these pseudo-labels and the ground-truth labels. Based on this process, the goal of optimizing the student network can be formulated as follows:
\begin{equation}
\mathcal{L}_{total} = \mathcal{L}_{sup} + \lambda_u \cdot \mathcal{L}_{unsup},
\end{equation}
where $\lambda_u$ is the hyper-parameter to adjust the contribution of the unsupervised loss.

For labeled data $D_{l} = \{S_i^l, Y_i^l\}_{i=1}^{N_l}$, the supervised loss $\mathcal{L}_{sup}$ for object detection consists of the classification loss $\mathcal{L}_{\mathrm{cls}}$, the regression loss $\mathcal{L}_{\mathrm{reg}}$~\cite{ren2015faster}.
\begin{equation}
\begin{aligned}
\mathcal{L}_{sup} = \sum_i \mathcal{L}_{cls}(S_i^{l}, {Y}_i^l) 
+ \mathcal{L}_{reg}(S_i^{l}, {Y}_i^l).
\end{aligned}
\end{equation}
The unsupervised loss $\mathcal{L}_{unsup}$ will be discussed in the chapter on Reliability-Guided Adaptive Constraint.

\subsection{Pseudo-Label Reliability Assessment (PLRA)}

In semi-supervised object detection, the performance enhancements of the student hinge on high quality pseudo-labels generated by the teacher. Current approaches often filter pseudo-labels based solely on a classification confidence threshold, neglecting the precision of bounding-box localization. Consequently, they end up introducing numerous boxes that are high confidence but coarse or even high confidence but incorrect as pseudo-labels.

As shown in Fig.~\ref{fig:Sonar}(b), we propose a pseudo-label reliability assessment scheme based on multi-view consistency. Specifically, for an unlabeled sonar image $S^u$, we first perform multi-view augmentation operations, including mirrored and scaling operations. The teacher network will predict the bounding boxes for \( S^u \) and its augmented view after augmentation, denoted as \( Y^{uo} \) and \( Y^{ua} \), respectively. Next, we match each bounding box in $Y^{uo}$ with the corresponding bounding box in $Y^{ua}$ according to their IoU values. The boxes in the bounding of the various augmented views that match those of the original view are denoted as $(y^{uo}, y^{ua})$.

First, for a given $\{y^{uo}_j\}_{j=1}^{N_j}$, we search for $\{y^{ua}_p\}_{p=1}^{N_p}$ in $Y^{ua}$ that has the highest IoU denoted as $IoU^*$ with $y^{uo}_j$. The computation is formulated as follows:
\begin{equation}
IoU^* = \mathop{\max}_{y_j^{uo} \in Y^{uo},\; y_p^{ua} \in Y^{ua}} \; \mathrm{IoU}(y_j^{uo},\; y_p^{ua}),
\end{equation}where \( N_j \) and \( N_p \) represent the number of pseudo-label boxes in the original view and the augmented view, respectively. After identifying the box \( y_p^{ua} \) that most closely matches the original view box \( y_j^{uo} \) in the augmented views, we comprehensively assess the reliability of the pseudo-labels generated by the teacher. To achieve this, we propose a reliability score \( R_j \) that jointly considers both the positional consistency and the category consistency of the predicted boxes across various augmented views. The calculation of the reliability score is as follows:
\begin{equation}
R_j = \mathrm{Sigmoid} \left(
    \frac{1}{H} \cdot
    \frac{
        \sum\limits_{a \in \mathcal{A}}
        \left[
            IoU^* \cdot \left( 1 - \left| s^o_j - s^a_{p} \right| \right) \cdot \beta
        \right]
    }
    {N_a}
\right),
\end{equation}
where $H$ is the reliability scaling factor, $\mathcal{A}$ denotes the set of augmented views, including mirrored and scaling operations. 
$s^o_j$ represents the class confidence score of the predicted box $y_j^{uo}$ from the original view, while $s_{p}^a$ denotes the class confidence of the predicted box $y_p^{ua}$. $N_a$ is the number of augmented views, which is set to 2, including mirrored and scaled views. The main purpose of $\beta$ is to determine whether the predicted class of the sonar image $S^u$ pseudo-label bounding box $y^{uo}_j$ is consistent with that of the corresponding augmented view pseudo-label bounding box $y_p^{ua}$. Its calculation formula is as follows:
\begin{equation}
\beta = 
\begin{cases}
1, & \text{if } c_j^o = c_p^a, \\
0, & \text{otherwise},
\end{cases}
\end{equation}
where $c_j^o$ denotes the class of the bounding box $y_j^{uo}$ and $c_p^a$ denotes the class of the bounding box $y_p^{ua}$. 

By comparing the teacher's predictions across different augmented views, we calculate a reliability score for each pseudo-label bounding box. Pseudo-labels that demonstrate higher consistency in predictions across different views will receive higher scores.


\subsection{Object Mixed Pseudo-Labels (OMPL)}

To fully utilize the high reliability pseudo-labels provided by the teacher and enhance the model's ability to recognize challenging samples (such as multiple small objects), we propose an object mixed pseudo-label strategy. This strategy involves rotating and scaling the pseudo-label bounding boxes with high reliability scores and mixing them with pseudo-labels from other samples to generate more pseudo-labels containing small and multiple objects. These pseudo-label bounding boxes can cover a substantial amount of irrelevant noise in the background of sonar images, thereby maximizing the utilization of valuable high reliability object pseudo-label bounding boxes.

As shown in Fig.~\ref{fig:Sonar}(c), we propose the object mixed pseudo-label strategy to fully leverage the high confidence pseudo-labels provided by the teacher. First, we obtain the set of pseudo-label bounding boxes $\widehat{y}^{u}_{S_k}= \left\{ y_{S_k}^{u} \;\middle|\; R_{j} > \hat{\gamma},\; k \in \{2, 3, \ldots, K\} \right\}$ with high reliability scores from the $k$-th image in the current batch. $\hat{\gamma}$ is the reliability score threshold, and $y^{u}_{S_k}$ denotes the set of pseudo-label bounding boxes for the $k$-th image in the current batch. $R_{j}$ represents the reliability score of the pseudo-label bounding boxes $y^{u}_{S_k}$. We can paste the augmented pseudo-label predicted boxes back onto the first image of the current batch. The operation can be described as follows:
\begin{equation}
S_1' = S_1 + \sum_{b_{S_k} \in \widehat{y}^{u}_{S_k}} \bar{F}_b \cdot {P}(S_1, b_{S_k}),
\end{equation}
where $S_1$ represents the first sonar image in the current batch and $S_1'$ represents the updated image after the paste operation is completed. $\widehat{y}^{u'}_{S_k}$ represents $ \widehat{y}^{u}_{S_k}$ after augmentation (the augmentation operations include random rotation and scaling of the pseudo-labels). ${P}(S_1, b_{S_k})$ represents pasting the pseudo-label boxes $b_{S_k}$ into the first image $S_1$. $b_{S_k}$ represents a bounding box in $ \widehat{y}^{u'}_{S_k}$. $\bar{F}_b $ is the object selection flag, which can be expressed as:
\begin{equation}
\bar{F}_b =
\begin{cases}
1, & \text{if } \mathrm{IoU}(b_{S_1}, b_{S_k}) < \delta, \\
0, & \text{otherwise}.
\end{cases}
\end{equation}
Only when the IoU between \( b_{S_k} \) and any pseudo-label box \( b_{S_1} \) in the first image \( S_1 \) is less than the IoU threshold \( \delta \) that we set will the paste operation be performed to avoid the pasted pseudo-label boxes overlapping.


\subsection{Reliability-Guided Adaptive Constraint}

 As shown in Fig.~\ref{fig:Sonar}(d), for the unlabeled data $D_{u} = \{S_i^u\}_{i=1}^{N_u}$, we utilize a reliability-guided approach to compute the unsupervised loss, adjusting the influence of each pseudo-label on the overall loss.
\subsubsection{Pseudo-Label Reliability Weighting (PLRW)}
We use reliability scores to differentiate the extent to which different pseudo-labels impact the loss function. We first evaluate the pseudo-label boxes provided by the teacher using the PLRA, obtaining a score for each pseudo-label box. The computed scores are then used to weight the pseudo-label boxes, so that boxes with higher reliability scores from the teacher are assigned greater weights when calculating the loss with the student. The loss $\mathcal{L}_{PLRW}$ guided by the reliability score is calculated as follows:
\begin{equation}
\mathcal{L}_{PLRW} = \frac{1}{N_j} \sum_{j=1}^{N_j} R_j \cdot (l_{cls} \left( b_j^{u}, T^{u} \right)
+ l_{reg} \left( b_j^{u}, T^{u} \right)),
\end{equation}
where ${T}^u$  denotes the set of pseudo-boxes generated by the teacher, $l_{cls}$ is the loss in box classification and $l_{reg}$ is the loss in box regression~\cite{xu2021end}. $N_j$ represents the number of candidate boxes predicted by the student. $b_j^{u}$ denotes the $j$-th predicted box in the set of candidate boxes predicted by the student, and $R_j$ denotes the reliability score of the pseudo-box of the teacher corresponding to the bounding box of the student.

\begin{figure*}[t]
    \centering
    \includegraphics[width=\linewidth]{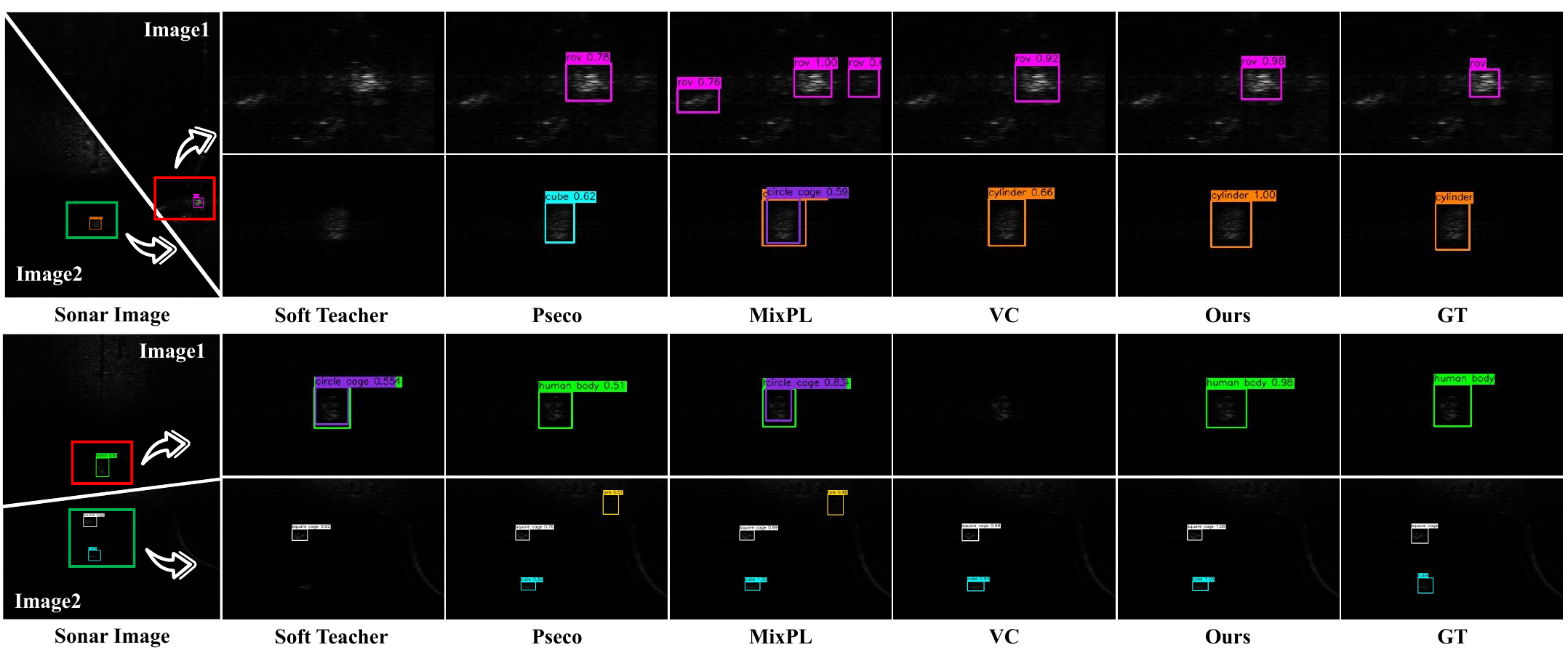}  
    \caption{Qualitative comparison of state-of-the-art object detection methods on the UATD dataset with 5\% labeled data.}
    
\label{fig:UATD1}
\end{figure*}

\subsubsection{Corner Point Regression Loss (CPRL)}
To further leverage pseudo-labels with high reliability scores from the teacher, we enhance the precision of the student’s bounding box localization by introducing a corner point regression loss. For each predicted bounding box $y_j^{u}\in {T}^u$ from the teacher, $R_j$ is the reliability score corresponding to $y_j^{u}$. If $R_j > \hat{\gamma}$, we apply corner point regression loss to strengthen the supervision of the teacher over the box location predictions of the student. The loss of corner point regression loss $\mathcal{L}_{CPRL}$ is as follows:
\begin{equation}
\mathcal{L}_{CPRL} = \frac{1}{N_j} \sum_{j}^{N_j} \frac{\left\| C^T_j - C^S_j \right\|_F^2}{l_j^2},
\end{equation}
where $N_j$ denotes the number of boxes that both exceed the teacher's reliability score threshold $\hat{\gamma}$ and match the predicted boxes of the student. $C^T_j$ denotes all corner points corresponding to the predicted boxes from the teacher, which can be expressed as a two-dimensional matrix, where each row denotes the 2D coordinates of a corner point~\cite{putra2025enhancing}. $C^S_j$ represents the corner coordinates of the predicted box from the student. ${\left\|\right\|_{F}^{2}}$ represents the Frobenius norm, which is the square root of the sum of the squares of all the elements in a matrix. $l_j^2$ represents the diagonal length of the minimum enclosing rectangle of the predicted matched boxes from the teacher and student. 

The final unsupervised loss $\mathcal{L}_{unsup}$ can be expressed as:
\begin{equation}
\mathcal{L}_{unsup} = \mathcal{L}_{PLRW} +\mathcal{L}_{CPRL}.
\end{equation}

\begin{table*}[t]   
  \label{tab:comparison}  
  \centering
  \small
  \begin{tabular*}{\linewidth}{@{\extracolsep{\fill}} c|ccc|ccc}  
    \toprule 
    \multirow{2}{*}{\diagbox[width=6cm]{\textbf{Method}}{\textbf{Dataset}}} & \multicolumn{3}{c|}{\textbf{UATD}} & \multicolumn{3}{c}{\textbf{FSOD}}\\
    & 1\% & 5\% & 10\% & 5\% & 10\% & 20\%\\
    \midrule
    Faster R-CNN~\cite{ren2015faster}(Supervised)    
          & 13.7 & 22.9 & 26.2 & 44.9 & 50.6 & 59.6\\
    \midrule 
    Soft Teacher~\cite{xu2021end}   
          & \underline{\textit{23.0(+9.3)}} & 30.8(+7.9) & 33.1(+6.9) & 56.4(+11.5) & 65.7(+15.1) & 72.2(+12.6)\\    
    Unbiased Teacher V2~\cite{liu2022unbiased}   
          & 21.6(+7.9) & 31.0(+8.1) & 33.4(+7.2) & 65.0(+20.1) & {70.1(+19.5)} & 72.8(+13.2)\\  
    Pseco~\cite{li2022pseco}   
          & 22.9(+9.2) & \underline{\textit{32.7(+9.8)}} & \underline{\textit{33.6(+7.4)}} & 65.7(+20.8) & \underline{\textit{70.5(+19.9)}} & \underline{\textit{74.6(+15.0)}}\\  
    MixPL~\cite{chen2023mixed}   
          & 22.2(+8.5) & 31.6(+8.7) & 32.2(+6.0) & 64.3(+19.4) & 70.2(+19.6) & 74.5(+14.9)\\  
    VC~\cite{chen2024virtual}   
          & 21.5(+7.8) & 30.9(+8.0) & 32.2(+6.0) & \underline{\textit{65.9(+21.0)}} & 69.6(+19.0) & 74.0(+14.4)\\    
    RSOD(Ours)   
          & {\textbf{23.2(+9.5)}} & {\textbf{34.7(+11.8)}} & {\textbf{34.8(+8.6)}} & {\textbf{67.4(+22.5)}} & {\textbf{72.6(+22.0)}} & {\textbf{75.5(+15.9)}}\\  
    \bottomrule  
  \end{tabular*}
  \caption{Quantitative comparisons on the UATD dataset and FSOD dataset under the mAP. Experiments on the UADT dataset were conducted under labeled data settings of 1\%, 5\%, and 10\%, while experiments on the FSOD dataset were conducted under labeled data settings of 5\%, 10\%, and 20\%. The best results are shown in \textbf{bold}, and the second best results are shown in \underline{\textit{italic}}.} 
  \label{tab:comp}
\end{table*}

\section{Experiment}
\subsection{Datasets and Experiment Settings}
\subsubsection{Dataset}
This study conducted comparison and ablation experiments on the FSOD dataset we collected and the publicly available UATD dataset~\cite{xie2022dataset}. The UATD dataset is widely used for object detection tasks in underwater sonar images~\cite{yang2024lightweight,wang2022mlffnet}. We utilized the original partition of the dataset, with 7,600 images in the training set and 800 images each in the test and validation sets. Since there is no current research on semi-supervised object detection in sonar images, we follow the protocols used for semi-supervised object detection~\cite{wang2023consistent}. In practice, we randomly selected 1\%, 5\%, and 10\% of the images in the training set as labeled data, while the remaining images were treated as unlabeled data.

For the FSOD dataset that we collected, we divided it into training, validation, and test sets in a 6:2:2 ratio. This dataset exhibits a long-tail distribution, with only 81 images belonging to the diver category. We randomly selected 5\%, 10\%, and 20\% of the images as labeled data, and the rest were considered unlabeled data.
\subsubsection{Evaluation Metrics}  
For all experiments, we evaluated the performance on the UATD and FSOD datasets, reporting the results using standard mean average precision (mAP) as an evaluation metric~\cite{xu2021end,chen2023mixed}.
\subsubsection{Implementation Details} 
To ensure a fair comparison, we follow STAC~\cite{sohn2020simple} to use FasterRCNN with FPN~\cite{lin2017feature} and ResNet-50 backbone~\cite{he2016deep} as our object detector, where the feature weights are initialized by the ImageNet-pretrained model, same as existing
works~\cite{DBLP:conf/iclr/LiuMHKCZWKV21,jeong2019consistency}. Furthermore, an asymmetric data augmentation strategy is applied to the unlabeled data, including weak augmentation by random flipping~\cite{li2022pseco} and strong augmentation through a greater degree of random flipping and random Gaussian blur~\cite{tang2021humble}. The entire model is trained for 180K iterations on a single NVIDIA RTX 4090 GPU. We use SGD optimizer for all detectors with a learning rate of 0.001, a momentum of 0.9, and a weight decay of 0.0001. 

\subsection{Main Results}

In this section, we compare our RSOD with two-stage classical fully supervised object detectors, Faster R-CNN~\cite{ren2015faster}. Furthermore, we compare it with various state-of-the-art semi-supervised object detection (SSOD) methods, including Unbiased Teacher V2~\cite{liu2022unbiased}, Soft Teacher~\cite{xu2021end}, MixPL~\cite{chen2023mixed}, Pseco~\cite{li2022pseco}, and VC~\cite{chen2024virtual}. Using publicly available source code, we have reimplemented these benchmarks on the proposed UATD and FSOD datasets.

\begin{table}[t]   
  \label{tab:comparison}  
  \centering
  \small
  \begin{tabular}{@{\hskip 5pt}c|ccc@{\hskip 5pt}}  
    \toprule 
    \multirow{2}{*}{\textbf{Method}} & \multicolumn{3}{c}{\textbf{UATD}} \\
    & 1\% & 5\% & 10\% \\
    \midrule
    Faster R-CNN    
          & 2.9 & 7.1 & 7.3 \\
    \midrule  
    Soft Teacher   
          & \underline{\textit{9.9(+7.0)}} & 11.2(+4.1) & 12.2(+4.9) \\    
    Unbiased Teacher V2   
          & {9.1(+6.2)} & 10.4(+3.3) & \underline{\textit{12.6(+5.3)}} \\  
    Pseco  
          & 8.9(+6.0) & 12.0(+4.9) & 12.3(+5.0) \\  
    MixPL   
          & 9.1(+6.2) & 10.1(+3.0) & 11.4(+4.1) \\  
    VC  
          & 5.7(+2.8) & \underline{\textit{12.2(+5.1)}} & 12.4(+5.1) \\    
    RSOD (Ours)   
          & {\textbf{10.1(+7.2)}} & \textbf{12.6(+5.5)} & \textbf{12.8(+5.5)} \\  
    \bottomrule  
  \end{tabular}
  \caption{Quantitative comparisons on the UATD dataset under the mAP(small) with different labeled data settings.}
  \label{tab:small}
\end{table}

\begin{table}[ht]   
\centering

\setlength{\tabcolsep}{5pt} 
\small
\resizebox{0.95\linewidth}{!}{
    \begin{tabular}{ccccccc}  
        \toprule  
        Model & EMA & PLRW & CPRL & OMPL & mAP  \\
        \midrule  
        M1 & - & - & - & - & 44.9  \\
        M2 & $\surd$ & - & - & - & 64.3  \\
        M3 & $\surd$ & $\surd$ & - & - & 65.2  \\
        M4 & $\surd$ & $\surd$ & $\surd$ & - & \underline{\textit{66.3}}  \\
        M5 & $\surd$ & $\surd$ & $\surd$ & $\surd$ & \textbf{67.4}  \\
        \bottomrule  
    \end{tabular}
} 
\caption{Ablation study on key components of RSOD conducted on the FSOD dataset with 5\% labeled data.} 
\label{tab:Ablation}  
\end{table}

\begin{figure}[t]
    \centering
    \includegraphics[width=\linewidth]{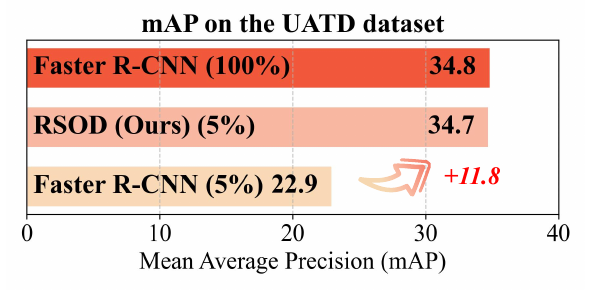}  
    \caption{mAP on the UATD dataset with 5\% labeled data.}
    
\label{fig:mAP}
\end{figure}
\begin{figure}[t]
    \centering
    \includegraphics[width=\linewidth]{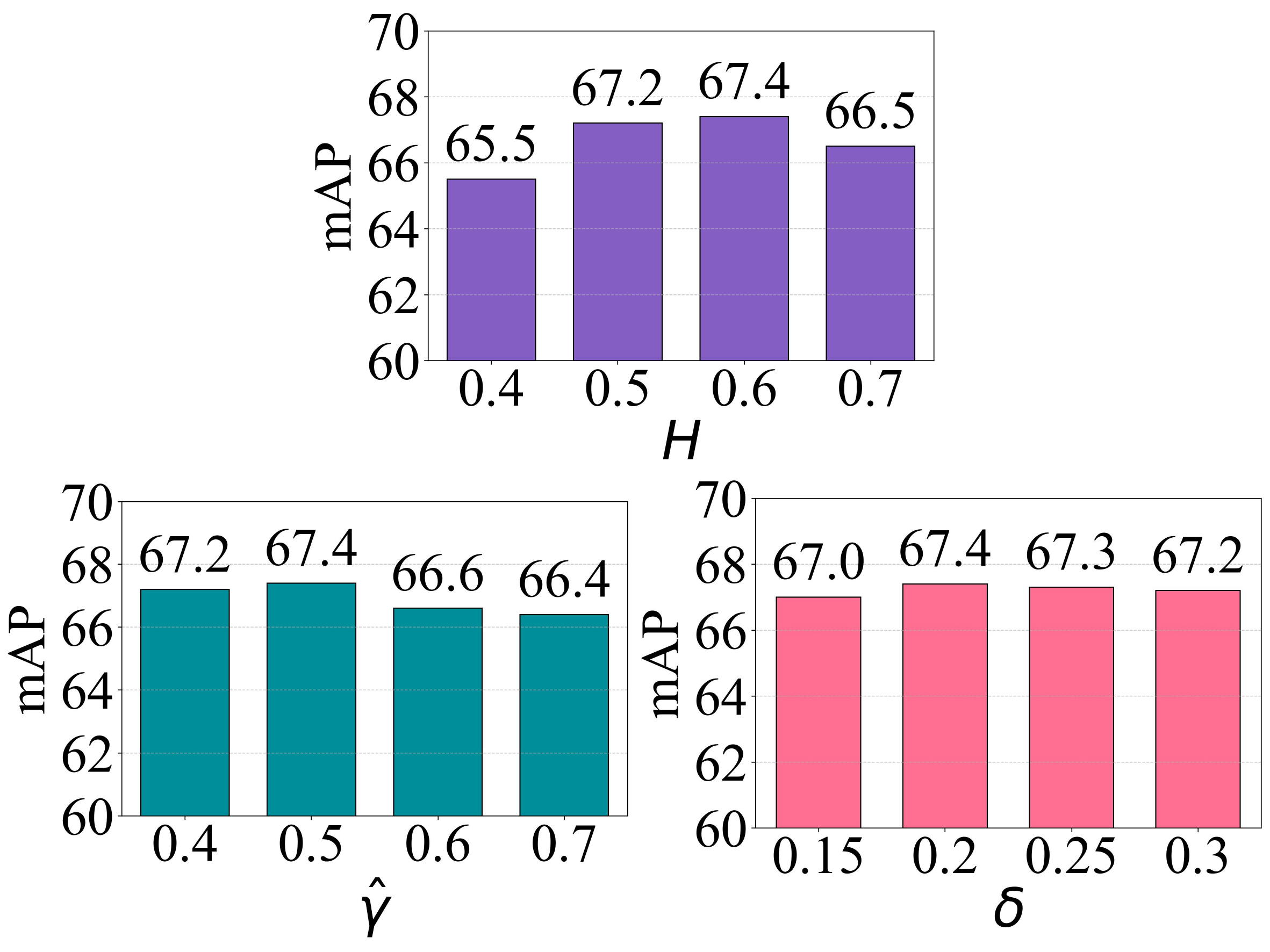}  
    \caption{mAP on the FSOD dataset with 5\% labeled data.}
    
\label{fig:parameter}
\end{figure}

\subsubsection{Qualitative Comparison}

As shown in Fig.~\ref{fig:UATD1}, in the detection of small objects in underwater sonar images, both the Soft Teacher and VC methods experience missed detections, while our approach successfully identifies the objects with high confidence. This indicates that our object mixed pseudo-label algorithm significantly enhances the detection ability for small objects. In contrast, both the Pseco and MixPL methods produce false detections, and their predicted bounding boxes have high confidence levels. Our method remains consistent with ground truth, highlighting that our reliability scoring assessment strategy has been effective in suppressing the influence of unstable pseudo-label boxes. Our method demonstrates overall performance that exceeds other approaches. Not only does it avoid missed detections and false detections, but it also provides high confidence for all correctly detected objects.

\subsubsection{Quantitative Comparison}

As shown in Table~\ref{tab:comp}, RSOD significantly outperforms other methods in the UATD and FSOD datasets, demonstrating the effectiveness of our proposed reliability score evaluation method and the object mixed pseudo-label algorithm. This approach notably increases the utilization of high reliability pseudo-labels while effectively suppressing the impact of unreliable pseudo-labels, successfully enhancing the predictive accuracy of the student. As shown in Table~\ref{tab:small}, RSOD also improves the ability of the model to predict small objects.

As shown in Fig.~\ref{fig:mAP}, on the UATD dataset, when the proportion of labeled data is only 5\%, the mAP of RSOD can compete against the fully supervised results obtained from the Faster R-CNN baseline detector using training data 100\%. This effectively alleviates the negative impact of scarce annotations on model performance.

\subsection{Ablation Study} \label{sec:ablation}
In this section, we conducted extensive research to validate our key
design and critical parameter settings on the FSOD dataset with 5\% labeled data.

\subsubsection{Component Analysis}
The results in Table~\ref{tab:Ablation} validate the effectiveness of key components in RSOD. M1 performed better with the Faster R-CNN detector that we used under fully supervised conditions. M2 indicates the results after introducing the teacher-student network. M3 shows the performance of the model after incorporating PLRW. M4 illustrates the model's performance following the incorporation of CPRL. M5 refers to the performance after applying the OMPL. From the experimental results, it can be seen that the introduction of the designed teacher-student network architecture has played a significant role. Guided by the reliability scores of the pseudo-label algorithm, both PLRW and CPRL effectively reduce the noise in the pseudo-label bounding boxes, greatly improving the mAP of the student. Our proposed object mixed pseudo-label strategy effectively enhances the model's detection capabilities for challenging samples, such as small objects and multiple objects.

\subsubsection{Parameter Analysis}
As shown in Fig.~\ref{fig:parameter}, we analyzed the key parameters during the experimental process, including the reliability scaling factor $H$, the reliability score threshold $\hat{\gamma}$, and the IoU threshold $\delta$. The experimental results indicate that setting $H$ to 0.6 effectively expands the gap in reliability scores between pseudo-labels, helping the model better distinguish between high-quality and low-quality pseudo-labels. Furthermore, when the reliability score threshold $\hat{\gamma}$ is set to 0.5 and the IoU threshold $\delta$ is set to 0.2, the model achieves the best result of 67.4. This indicates that the object mixed pseudo-label strategy and corner point regression loss assist the student detector in achieving optimal performance.

\section{Conclusion}
In this paper, we combine the teacher-student architecture with pseudo-label reliability assessment methods to effectively utilize unlabeled data. The student is better able to distinguish and learn different pseudo-labels with the support of reliability scores, significantly mitigating the negative impact of noisy pseudo-labels. Additionally, the object mixed pseudo-label strategy effectively enhances the model's ability to handle difficult samples, such as small and multiple targets. We conducted extensive experiments on two datasets, and the results demonstrate that RSOD exhibits significant superiority over other methods in various scenarios with limited labels.

\section{Acknowledgments}
This work was supported by the National Natural Science Foundation of China under Grant Nos. 62495085, 62027826, 62442603,
62272083, 62472066 and 62495083 and by the Key Research and Development Program of Liaoning Province,
under No. 2023JH26/10200014.

\bibliography{aaai2026}

\end{document}